\documentclass[10pt,twocolumn,letterpaper]{article}

\usepackage{cvpr}
\usepackage{times}
\usepackage{epsfig}
\usepackage{graphicx}
\usepackage{amsmath}
\usepackage{amssymb}

\usepackage{ dsfont }
\usepackage{cuted}
\usepackage{capt-of}
\usepackage{verbatim} 

\usepackage[pagebackref=true,breaklinks=true,letterpaper=true,colorlinks,bookmarks=false]{hyperref}

 \cvprfinalcopy 

\def\cvprPaperID{4558} 

\ifcvprfinal\fi
\begin{document}

\title{Combining 3D Morphable Models: A Large scale Face-and-Head Model}

\author{Stylianos Ploumpis$^{1,3}$ 
        \hspace{0.8cm}
        Haoyang Wang$^1$
        \hspace{0.8cm}
        Nick Pears$^2$\\
        William A. P. Smith$^2$ 
        \hspace{0.8cm}
        Stefanos Zafeiriou$^{1,3}$ \and
$^1$Imperial College London, UK
\hspace{0.8cm}
$^2$University of York, UK
\hspace{0.8cm}
$^3$FaceSoft.io
\\
{\tt\footnotesize $^1$\{s.ploumpis,haoyang.wang15,s.zafeiriou\}@imperial.ac.uk}
\hspace{0.5cm}
{\tt\footnotesize $^2$\{nick.pears,william.smith\}@york.ac.uk}
}

\maketitle

\begin{strip}\centering
\includegraphics[width=1\textwidth]{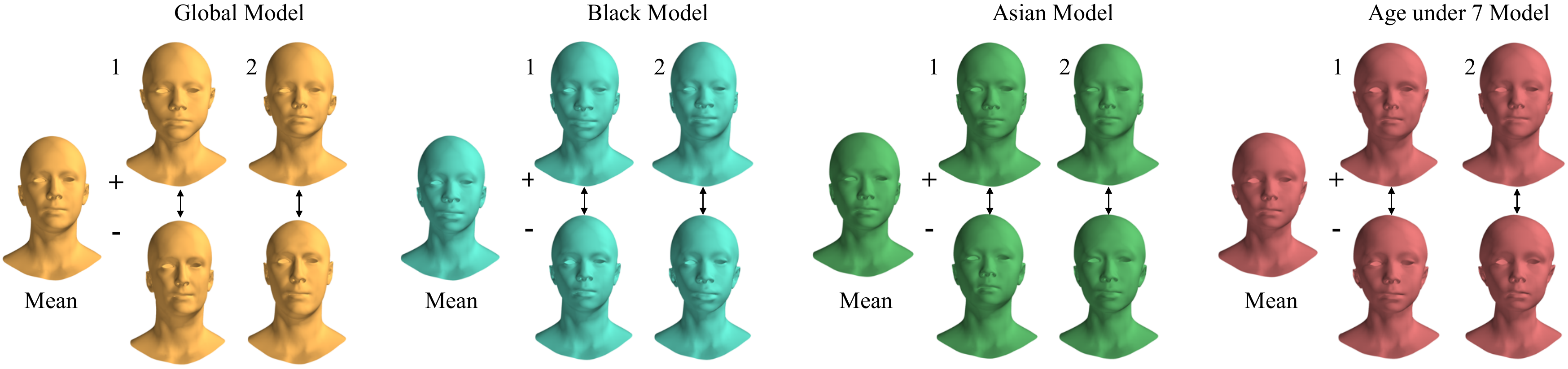}
\captionof{figure}{\small{The bespoke Combined Face \& Head Models. Visualisation of the first two shape components along with the mean head shape.}
\label{fig:feature-graphic}}
\end{strip}

\begin{abstract}
Three-dimensional Morphable Models (3DMMs) are powerful statistical tools for representing the 3D surfaces of an object class. In this context, we identify an interesting question that has previously not received research attention: is it possible to combine two or more 3DMMs that (a) are built using different templates that perhaps only partly overlap, (b) have different representation capabilities and (c) are built from different datasets that may not be publicly-available? In answering this question, we make two contributions. First, we propose two methods for solving this problem: i.~use a regressor to complete missing parts of one model using the other, ii.~use the Gaussian Process framework to blend covariance matrices from multiple models. Second, as an example application of our approach, we build a new face-and-head shape model that combines the variability and facial detail of the LSFM with the full head modelling of the LYHM. The resulting combined shape model achieves state-of-the-art performance and outperforms existing head models by a large margin. Finally, as an application experiment, we reconstruct full head representations from single, unconstrained images by utilizing our proposed large-scale model in conjunction with the FaceWarehouse blendshapes for handling expressions.

\end{abstract}

\section{Introduction}

Due to their ability of inferring and representing 3D surfaces, 3D Morphable Models (3DMMs) have many applications in computer vision, computer graphics, biometrics, and medical imaging \cite{blanz2003face,hu2016face,aldrian2013inverse,staal2015describing}. Many registered raw 3D images (`scans') are required for correctly training a 3DMM, which comes at a very large cost of manual labour for collecting and annotating such images with meta data. Sometimes, only the resulting 3DMMs become available to the research community, and not the raw 3D images. This is particularly true of 3D images of the human face/head, due to increasingly stringent data protection regulations. Furthermore, even if 3DMMs have overlapping parts, their resolution and ability to express detailed shape variation may be quite different, and we may wish to capture the best properties of multiple 3DMMs within a single model. However, it is currently extremely difficult to combine and enrich existing 3DMMs with different attributes that describe distinct parts of an object without such raw data. Therefore, in this paper,  we present a general approach that can be employed to combine 3DMMs from different parts of an object class into a single 3DMM. Due to their widespread use in the computer vision community, we fuse 3DMMs of the human face and the full human head, as our exemplar, thus creating the first \emph{combined, large-scale, full-head} morphable model. The technique is readily extensible to incorporate detailed models of the ear \cite{dai2018ear} and the body, and indeed is applicable to any object class well-described by 3DMMs.

More specifically, although there have been many models of the human face both in terms of identity \cite{huber2016multiresolution,zhu2015discriminative,zhu2016face} and expression \cite{bronstein2003expression,zhu2015high}, very few deal with the complete head anatomy \cite{dai20173d}. Building a high-quality, large-scale statistical model that describes the anatomy of the full human head paves directions across numerous disciplines. First, it will assist craniofacial clinicians in diagnosis, surgical planning, and assessment. Second, generating proportionally correct head models based on the geometry of the face will aid computer graphics designers to create realistic avatar-like representations. Finally, a head model will give opportunities that aim at reconstructing a full head representation from data-deficient sources, such as 2D images.

Our key contributions are: (i) a methodology that aims to fuse shape-based 3DMMs, using the human face and head as an exemplar. (Note that the texture component of the 3DMM is out-of-scope of this paper and the subject of our future work.) In particular, we propose both a regression method based on latent shape parameters, and a covariance combination approach, utilized in a Gaussian process framework, (ii) a combined large-scale statistical model of the human head in terms of ethnicity, age and gender that is significantly more accurate than any other existing head morphable model - we make this publicly-available for the benefit of the research community, including versions with and without eyes and teeth, and (iii) an application experiment in which we utilize the combined 3DMM to perform full head reconstruction from unconstrained single images, also utilizing the FaceWarehouse blendshapes to handle facial expressions. 

\section{Face and head model literature}
\label{sec:faces}
The first 3DMM was proposed by Blanz and Vetter \cite{blanz1999morphable}. They were the first to to recognize the generative capabilities of a 3DMM and they proposed a
technique to capture the variations of 3D faces. Only 200 scans were used to build the model (100 male and 100 female) where dense correspondences were computed based on optical flow that depends on an energy function that describes both the shape and texture. The Basel Face Model (BFM) is the most widely-used and well-known 3DMM, which was built by Paysan \etal \cite{paysan20093d} and utilizes a better registration method than the original Blanz-Vetter 3DMM. They use a known template mesh in which all the vertices have known positions and then they register it to the training scans by utilizing an optimal step Non-rigid Iterative Closest Point algorithm (NICP) \cite{amberg2007optimal}. Standard PCA was employed as a dimensionality reduction technique to construct their model.

Recently, Booth \etal \cite{booth20163d} built a Large-scale Face Model (LSFM) by utilizing nearly $10,000$ face scans. The model is constructed by applying a weighted version of the optimal-step NICP algorithm \cite{de2010optimal}, followed by a Generalized Procrustes Analysis (GPA) and standard PCA. Due to the large number of facial scans, a robust automated procedure was carried out including 3D landmark localization and error pruning of badly registered scans. This work was the first to introduce bespoke models in terms of age, gender and ethnicity, and is the most information-rich 3DMM of face shapes in neutral expression produced to date.


 Li et al  \cite{li2017learning} used a total of $3,800$ head scans from the US and European CEASAR body scan database \cite{robinette2002civilian} to build a statistical model of the entire head. The aim of this work focuses mainly on the temporal registration of 3D scans rather than on the topology of the head area. The data consists of full body scans and the resolution in which the head topology was recorded in is insufficient to depict correctly the shape of each individual human head. In addition, the template used for registration in this method is extremely sparse with only $5,000$ vertices which makes it difficult to accurately represent the entire head. Moreover, the registration process incorporates coupling weights for the back of head and the back of the neck, which constrains drastically the actual statistical variation of the entire head area. An extension of this work is proposed in \cite{ranjan2018generating} in which a non-linear model is constructed using convolution mesh autoencoders focusing on facial expressions, but still it lacks the statistical variation of the full cranium.
  Similarly, in the work of Hu and Saito \cite{hu2017avatar}, a full head model is created from single images mainly for real-time rendering. The work aims at creating a realistic avatar model which includes 3D hair estimation. The head topology is considered to be unchanged for all subjects and only the face part of the head is a statistically-correct representation.
 
The most accurate craniofacial 3DMM of the human head both in terms of shape and texture, is the Liverpool-York Head model (LYHM) \cite{dai20173d}. In this work, global craniofacial 3DMMs and demographic sub-population 3DMMs were built from 1,212 distinct identities. 
Although this work is the first that describes the statistical correlation between the cranium and the face part, it lacks detail of the facial characteristics, as the spatial resolution of the facial region is not significantly higher than the cranial region. In effect, the variance of the cranial and neck areas dominates that of the facial region in the PCA parameterization. Also, although the model describes how the cranium is affected given the age of the subject, it is biased in terms of ethnicity, due to the lack of ethnic diversity in the dataset.
\section{Face and head shape combination}
\label{sec:face_and_head}
In this section, we propose two methods to combine the LSFM face model with the LYHM full head model. The first approach, utilizes the latent PCA parameters and solves a linear least squares problem to approximate the full head shape, whereas the second constructs a combined covariance matrix that is later utilized as a kernel in a Gaussian Process Morphable Model (GPMM) \cite{luthi2017gaussian}.

\subsection{Regression modelling}
\label{reg_combination}
\begin{figure*}[h]
\begin{center}
\includegraphics[width=1\linewidth]{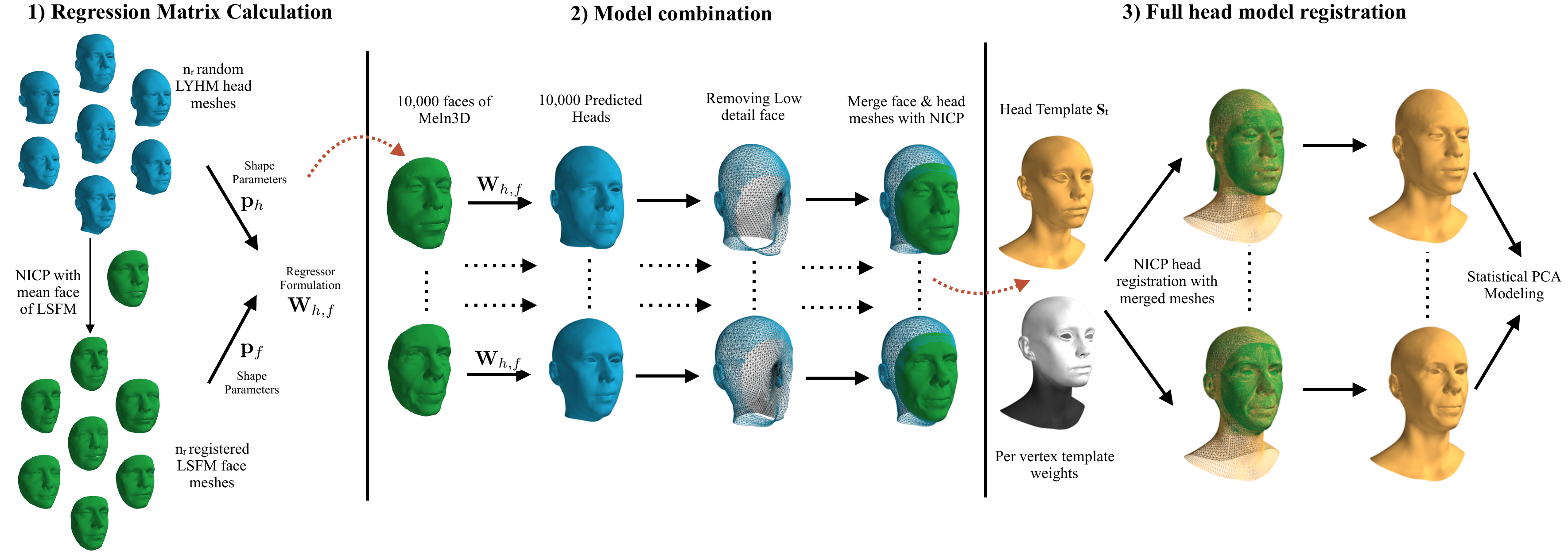}
\caption{The regression modeling pipeline. 1) The left part illustrates the  matrix formulation from the original LYHM head model; 2) the central part demonstrates how we utilize the \emph{MeIn3D} database to produce highly-detailed head shapes; 3) the final part on the right depicts the registration framework along with the per-vertex template weights and the statistical modeling.}
\label{fig:reg_diagram}
\end{center}
\end{figure*}

Figure \ref{fig:reg_diagram} illustrates the three-stage regression modeling pipeline, which comprises 1) regression matrix calculation, 2) model combination and 3) full head model registration followed by PCA modeling. Each stage is now described.

For stage 1, let us denote the 3D mesh (shape) of an object with $N$ points as a $3N \times 1$ vector
\begin{equation}
    \mathbf{S} = [\mathbf{x}_1^T\dots\mathbf{x}_N^T]^T = [x_1, y_1, z_1,\dots x_N, y_N, z_N]^T
\end{equation}
The LYHM is a PCA generative head model with $N_h$ points, described by an orthonormal basis after keeping the first $n_h$ principal components $\mathbf{U}_{h} \in \mathds{R}^{3N_h\times n_h} $ and the associated $\mathbf{\lambda}_{h}$ eigenvalues. This model can be used to generate novel 3D head instances as follows:
\begin{equation}
\mathcal{\mathbf{S}}_{h}(\mathbf{p}_{h}) = {\mathbf{m}}_{h} + {\mathbf{U}}_{h} \mathbf{p}_{h}
\label{equ:head_shape_instance}
\end{equation}
where $ \mathbf{p}_{h} = \left[p_{h_1} \ldots p_{h_{n_h}} \right ]^T$ are the $n_h$ shape parameters. Similarly the LSFM face model with $N_f$ number of points, is described by a corresponding orthonormal basis after keeping the $n_f$ principal components $\mathbf{U}_f \in \mathds{R}^{3N_f\times n_f} $ and the associated $\mathbf{\lambda}_f$ eigenvalues. The model generates novel 3D faces instances by:
\begin{equation}
\mathcal{\mathbf{S}}_{f}(\mathbf{p}_{f}) = {\mathbf{m}}_{f} + {\mathbf{U}}_{f} \mathbf{p}_{f}
\label{equ:face_shape_instance}
\end{equation}
where $ \mathbf{p}_{f} = \left[p_{f_1} \ldots p_{f_{n_f}} \right ]^T$ are the $n_f$ shape parameters. 

In order to combine the two models, we synthesize data directly from the latent eigenspace of the head model ($\mathbf{U}_{h}$) by drawing random samples from a Gaussian distribution defined by the principal eigenvalues of the head model. The standard deviation for each of the distributions is equal to the square root of the eigenvalue. In that way we produce randomly $n_r$ distinct shape parameters.


After generating the random full head instances we apply non-rigid registration (NICP) \cite{de2010optimal} between the head meshes and the cropped mean face of the LSFM face model. We perform this task in each one of the $n_r$ meshes in order to get the facial part of the full head instance and describe it in terms of the LSFM topology. Once we acquire those registered meshes we project them to the LSFM subspace and we retrieve the corresponding shape parameters. Thus, for each one of the randomly produced head instances, we have a pair of shape parameters ($\mathbf{p}_{h}, \mathbf{p}_{f}$) corresponding to the full head representation and to the facial area respectively.

By utilizing those pairs we construct a matrix $\mathbf{C}_h \in\mathds{R}^{n_h\times n_r }$ where we stack all the head shape parameters and a matrix $\mathbf{C}_f \in\mathds{R}^{n_f\times n_r }$ where we stack the face shape parameters from the LSFM model. We would like to find a matrix $\mathbf{W}_{h,f} \in \mathds{R}^{n_h\times n_f } $ to describe the mapping from the LSFM face shape parameters $\mathbf{p}_{f}$ to the corresponding LYHM full head shape parameters $\mathbf{p}_{h}$. We solve this by formulating a linear least square problem that minimizes:
\begin{equation}
 \left\lVert \mathbf{C}_h - \mathbf{W}_{h,f} \mathbf{C}_f \right\rVert^2
\label{equ:normal_equ_head_face}
\end{equation}
By utilizing the normal equation, the solution of Eq. \ref{equ:normal_equ_head_face} is readily given by:
\begin{equation}
\mathbf{W}_{h,f} = \mathbf{C}_h\mathbf{C}_f^T \left( \mathbf{C}_f\mathbf{C}_f^T \right)^{-1}
\end{equation}
where $\mathbf{C}_f^T \left( \mathbf{C}_f\mathbf{C}_f^T \right)^{-1}$ is the right pseudo-inverse of $\mathbf{C}_f$. Given a 3D face instance $\mathbf{S}_f$, we derive the 3D shape of the full head, $\mathbf{S}_h$, as follows:
\begin{equation}
\mathcal{\mathbf{S}}_h ={\mathbf{m}}_{h} + \mathbf{U}_{h} \mathbf{W}_{h,f} \mathbf{U}_{f}^T\left( \mathbf{S}_f - {\mathbf{m}}_{f}\right)
\end{equation}
In this way we can map and predict the shape of the cranium region for any given face shape in terms of LYHM topology. 

In stage 2 (Fig.~\ref{fig:reg_diagram}), we employ the large MeIn3D database \cite{booth20163d} which includes nearly $10,000$ 3D face images, and we utilize the $\mathbf{W}_{h,f}$ regression matrix to construct new full head shapes that we later combine with the real facial scans. We achieve this by discarding the facial region of the the full head instance which has less detailed information and we replace it with the registered LSFM face of the MeIn3D scan. In order to create a unique instance we merge the meshes together by applying a NICP framework, where we deform only the outer parts of the facial mesh to match with the cranium angle and shape so that the result is a smooth combination of the two meshes. Following the formulation in \cite{de2010optimal}, this is accomplished by introducing higher stiffness weights in the inner mesh (lower on the outside) while we apply the NICP algorithm. To compute those weights we measure the Euclidean distance of a given point from the nose tip of the mesh and we assign a relative weight to that point. The bigger the distance from the nose tip, the smaller the weight of the point.

One of the drawbacks of the LYHM is the arbitrary neck circumference, where the neck tends to get broader when the general shape of the head increases. In stage 3 (Fig. \ref{fig:reg_diagram}), we aim at excluding this factor from our final head model by applying a final NICP step between the merged meshes and our head template $\mathbf{S}_t$. We utilized the same framework as before with the point-weighted strategy where we assign weights to the points based on their Euclidean distance from the center of the head mass. This helps us avoid any inconsistencies of the neck area that might appear from the regression scheme. For the area around the ear, we have introduced 50 additional landmarks to control the registration and preserve the general shape of the ear area.

After implementing the aforementioned pipeline for each one of the $10,000$ meshes, we perform PCA on the points of the mesh and we acquire a new generative full head model that exhibits more detail in the face area in combination with bespoke head shapes.

\subsection{Gaussian process modeling}
Gaussian processes for model combination is a less complicated and more robust technique that does not generate irregular head shapes due to poor regression values.

The concept of Gaussian Process Morphable Models (GPMMs) was recently introduced in \cite{luthi2017gaussian,gerig2018morphable,koppen2018gaussian}. The main contribution of GPMMs is the generalization of classic Point Distribution Models (such as are constructed using PCA), with the help of Gaussian processes. A shape is modeled as a deformation $u$ from the reference shape $\mathbf{S}_R$ i.e. a shape can be represented as: 
\begin{equation}
    \mathbf{S} = \lbrace \mathbf{x} + u(\mathbf{x}) | \mathbf{x}\in \mathbf{S}_R \rbrace
\end{equation} 
where $u$ is a deformation function $u : \Omega \rightarrow \mathds{R}^{3}$ with $\Omega \supseteq  \mathbf{S}_R$. The deformations are modeled as a Gaussian process $u \sim \mathcal{GP}\left(\mu, k\right)$. Where $\mu : \Omega \rightarrow \mathds{R}^{3} $ is the mean deformation and $k : \Omega \times \Omega  \rightarrow \mathds{R}^{3\times3}$ is a covariance function or kernel. 

The Gaussian process model is capable of operation outside of the space of valid face shapes. This depends highly on the kernels chosen for this task. In the classic approaches, the deformation function is learned through a series of typical example surfaces $\mathbf{S}_1, \ldots,\mathbf{S}_n$ where a set of deformation fields is learned $\lbrace u_1,\ldots, u_n\rbrace, u_i(\mathbf{x})  : \Omega \rightarrow \mathds{R}^{d}$ where $u_i(\mathbf{x})$ denotes the deformation field that maps a point $\mathbf{x}$ on the reference shape to the corresponding point on the $i_{th}$-training surface. 

A Gaussian process $\mathcal{GP}\left( \mu_{PDM},k_{PDM}\right)$ that models this characteristic deformations is obtained by estimating the empirical mean: 
\begin{equation}
    \mu_{PDM}\left(\mathbf{x}\right) = \frac{1}{n}\sum_{i=1}^n u_{i}(\mathbf{x})
\end{equation}
and the covariance function:
\begin{equation}
\begin{split}
    k_{PDM}\left( \mathbf{x}, \mathbf{y}\right) = \frac{1}{1-n}\sum_{i=1}^n \left( u_{i}(\mathbf{x}) - \mu_{PDM}(\mathbf{x}) \right) \\
    \left( u_{i}(\mathbf{y}) - \mu_{PDM}(\mathbf{y}) \right)^T
\end{split}
\end{equation}
This kernel is defined as the empirical/sample covariance kernel. This specific Gaussian process model is a continuous analog to a PCA model and it operates in the facial deformation spectrum. 
For each one of the models (LYHM, LSFM), we know the principal orthonormal basis and the eigenvalues. Hence the covariance matrix for each model is defined:
\begin{equation} \label{equ:compute_covariance}
\begin{split}
    \mathbf{K}_h = \mathbf{U}_h \mathbf{ \Lambda}_h  \mathbf{U}_h^T \\
    \mathbf{K}_f = \mathbf{U}_f \mathbf{ \Lambda}_f  \mathbf{U}_f^T 
\end{split}
\end{equation}
where $\mathbf{K}_h\in \mathds{R}^{3N_h\times 3N_h}$ and $\mathbf{K}_f\in \mathds{R}^{3N_f\times 3N_f}$ are the covariance matrices, and the $\mathbf{\Lambda}_h\in \mathds{R}^{3n_h\times 3n_h}$ and $\mathbf{\Lambda}_f\in \mathds{R}^{n_f\times n_f}$ are diagonal matrices with the eigenvalues in their the main diagonal of the head and face model respectively.

We aim at constructing a universal covariance matrix $\mathbf{K}_{U} \in \mathds{R}^{3N_U\times 3N_U}$ that accommodates the high detailed facial properties of the LSFM and the head distribution from the LYHM. We keep, as a reference, the mean of the head model and we non-rigidly register the mean face of the LSFM. Both PCA models must be in the same scale space for this method to work, which was not necessary for the regression method. Similarly, we register our head template $\mathbf{S}_t$ by utilizing the same pipeline as before for full head registration, which is going to be used as the reference mesh for the new joined covariance matrix.


For each point pair $i,j$ in $\mathbf{S}_t$, there exists a local covariance matrix $\mathbf{K}_U^{i,j} \in \mathds{R}^{3\times3}$. In order to calculate its value, we begin by projecting the points onto the mean head mesh. If both points lie outside the face area that the registered mean mesh of LSFM covers, we identify their exact location in the mean head mesh in terms of barycentric coordinates $(c_1^i, c_2^i, c_3^i)$ for the $i_{th}$ point and $(c_1^j, c_2^j, c_3^j)$ for the $j_{th}$ point with respect to their corresponding triangles $\mathbf{t}_i = [\mathbf{v}_1^T,\mathbf{v}_2^T,\mathbf{v}_3^T]^T,\mathbf{t}_j = [\mathbf{k}_1^T,\mathbf{k}_2^T,\mathbf{k}_3^T]^T$.

\begin{figure}[h]
\begin{center}
\includegraphics[width=1\linewidth]{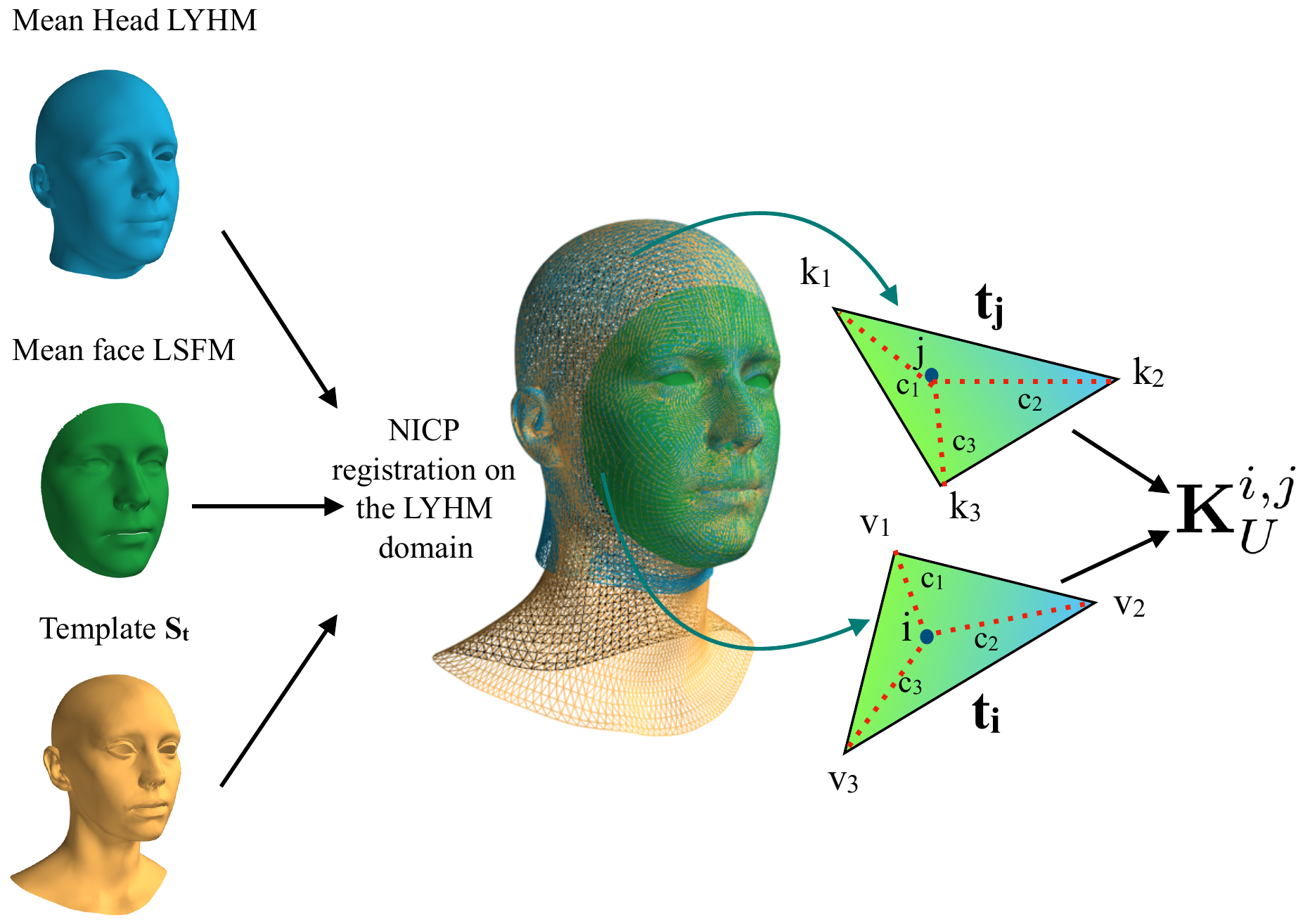}
\caption{A graphical representation of the non-rigid registration of all mean meshes along with our head template $\mathbf{S}_t$ and the calculation of the local covariance matrix $\mathbf{K}_U^{i,j}$ based on the locations of the $i_{th}$ and $j_{th}$ points.}
\label{fig:triangles}
\end{center}
\end{figure}

Each vertex pair $(v,k)$ in between the triangles, has an individual covariance matrix $\mathbf{K}_h^{v,k} \in \mathds{R}^{3\times3}$ with $\mathbf{K}_h^{v,k}\supseteq\mathbf{K}_h$. Therefore, we blend those local vertex-covariance matrices to acquire our final local $\mathbf{K}_U^{i,j}$ as follows:
\begin{equation}
\begin{split}
    \mathbf{K}_U^{i,j} =  \frac{\sum_{v=1}^3\sum_{k=1}^3 w_{v,k}^{i,j} \mathbf{K}_h^{v,k}}{\sum_{v=1}^3\sum_{k=1}^3 w_{v,k}^{i,j}}
\end{split}
\end{equation}
where $w_{v,k}^{i,j} = \frac{c_v^i + c_k^j}{2}$ is a weighting scheme based on the barycentric coordinates of the $(i,j)$ points. An illustration of the aforementioned methodology can be seen in Figure~\ref{fig:triangles}.

\begin{figure*}[h]
\begin{center}
\includegraphics[width=1\linewidth]{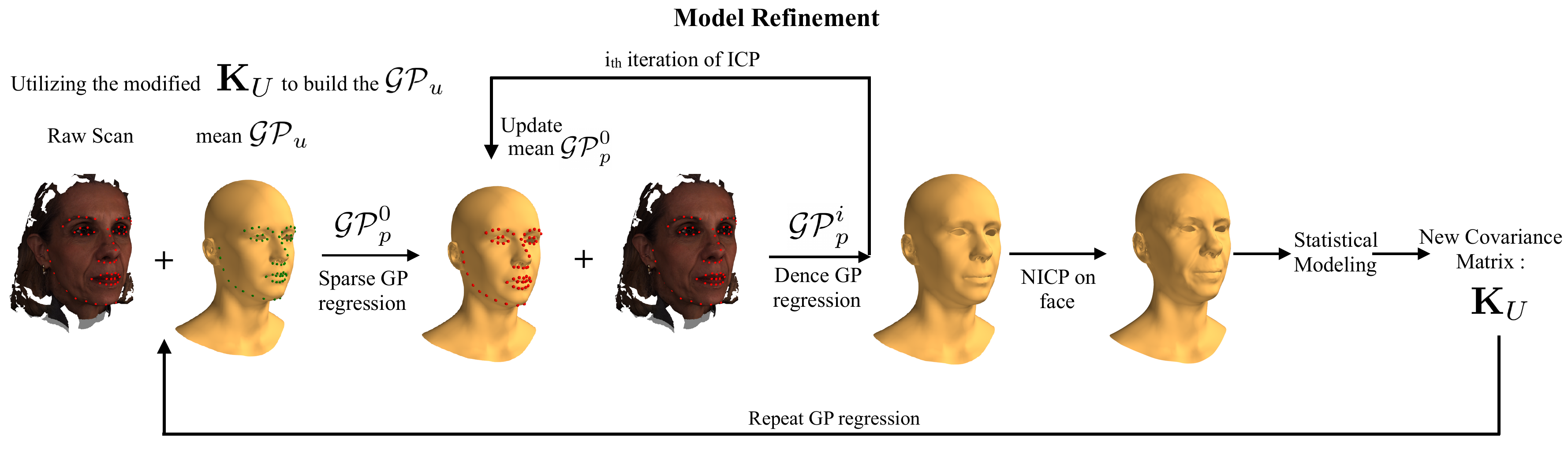}
\caption{The model refinement pipeline. We start with the GP model defined by the universal covariance matrix. For each scan in the \emph{MeIn3D} dataset we obtain full head reconstruction with GP Regression using the sparse landmarks and dense ICP algorithm. We then non-rigidly align the face region of the full head reconstruction to the scan, and build a new sample covariance matrix to update our model.}
\label{fig:intrinsic_exp}
\end{center}
\end{figure*}

In the case where the points lie in the face area, we initially repeat the same procedure by projecting and calculating a blended covariance matrix $\mathbf{K}_f^{i,j}$ given the mean face mesh of LSFM, followed by a blended covariance matrix $\mathbf{K}_h^{i,j}$ calculated given the mean head mesh of LYHM. We formulate the final local covariance matrix as:
\begin{equation}
    \mathbf{K}_U^{i,j} = \rho_{i,j}\mathbf{K}_h^{i,j} + (1-\rho_{i,j})\mathbf{K}_f^{i,j}
\end{equation}
where $\rho_{i,j} = \frac{\rho_i + \rho_j}{2}$ is a normalized weight, based on the Euclidean distances $(\rho_i,\rho_j)$ of the $(i,j)$ points from the nose-tip of the registered meshes. We apply this weighting scheme to smoothly blend the properties of the head and face model and to avoid the discontinuities that appear on the borders of the face and head area. 

Lastly, when the points belong to different areas \ie ($i_{th}$ point on face, $j_{th}$ point on head) we simply follow the first method that exploits just the head covariance matrix $\mathbf{K}_h$, since the correlation of the face/head shape only exist in the LYHM. After repeating the aforementioned methodology for every point pair in $\mathbf{S}_t$ and calculating the entire joined covariance matrix $\mathbf{K}_{U}$, we are able to sample new instances from the Gaussian process morphable model.

\subsection{Model Refinement}
To refine our model, we begin by exploiting the already trained GPMM of the previous section. 
With our head template $\mathbf{S}_{t}$ and the universal covariance matrix $\mathbf{K}_{U}$, we define a kernel function:
\begin{equation}
    k_{U}(\mathbf{x}, \mathbf{y}) = \mathbf{K}_{U}^{CP(\mathbf{S}_{t}, \mathbf{x}), CP(\mathbf{S}_{t}, \mathbf{y})}
\end{equation}
where $\mathbf{x}$ and $\mathbf{y}$ are two given points from the domain where the Gaussian process is defined and the function $CP(\mathbf{S}_{t}, \mathbf{x})$ returns the index of the closest point of $\mathbf{x}$ on the surface $\mathbf{S}_{t}$. We then define our GPMM as:
\begin{equation}
    \mathcal{GP}_{U}(\mu_{U}, k_{U})
\end{equation}
where $\mu_{U}(\mathbf{x}) = [0, 0, 0]^{T}$. For each scan in the MeIn3D dataset, we first try to reconstruct a full head registration with our GPMM using Gaussian Process Regression \cite{luthi2017gaussian, gerig2018morphable}. Given a set of observed deformations $\mathbf{X}$ subject to Gaussian noise $\epsilon \sim \mathcal{N}(0, \sigma^{2})$, Gaussian process regression computes a posterior model $\mathcal{GP}_{p}(\mu_{p}, k_{p}) = posterior(\mathcal{GP}, \mathbf{X})$. The landmark pairs between a reference mesh and the raw scan define a set of sparse mappings, which tells us exactly how the points on the reference mesh will deform. Any sample from this posterior model will then have fixed deformations on our observed points \ie facial landmarks. The mean $\mu_{p}$ and covariance $k_{p}$ are computed as:
\begin{equation}
    \mu_{p}(\mathbf{x}) = \mu(\mathbf{x}) + K_{\mathbf{X}}(\mathbf{x})^{T}(\mathbf{K}_{\mathbf{XX}} + \sigma^{2}\mathbf{I})^{-1}\mathbf{X}
\end{equation}
\begin{equation}
    k_{p}(\mathbf{x}, \mathbf{y}) = k_{u}(\mathbf{x}, \mathbf{y}) - K_{\mathbf{X}}(\mathbf{x})^{T}(\mathbf{K}_{\mathbf{XX}} + \sigma^{2}\mathbf{I})^{-1}K_{\mathbf{X}}(\mathbf{y})
\end{equation}
where 
\begin{equation}
K_{\mathbf{X}}(\mathbf{x}) = (k_{U}(\mathbf{x}, \mathbf{x}_{i})), \forall\; \mathbf{x}_{i} \in \mathbf{X}
\end{equation}
\begin{equation}
    \mathbf{K}_{\mathbf{XX}} = (k_{U}(\mathbf{x}_{i}, \mathbf{x}_{j})), \forall\;    \mathbf{x}_{i}, \mathbf{x}_{j} \in \mathbf{X}
\end{equation}

For a scan $\mathbf{S}$ with landmarks $\mathbf{L}_{\mathbf{S}} = \{\mathbf{l}_{1}, ... \mathbf{l}{n}\}$, we first compute a posterior model based on the sparse deformations defined by the landmarks:
\begin{equation}
    \mathcal{GP}_{p}^{0}(\mu_{p}^{0}, k_{p}^{0}) = posterior(\mathcal{GP}_{U}, \mathbf{L}_{\mathbf{S}} - \mathbf{L}_{\mathbf{S}_{t}})
\end{equation}
We then refine the posterior model with Iterative Closest Point algorithm. More specifically, at each iteration $i$ we compute the current regression result as $\mathbf{S}_{reg}^{i} = \{\mathbf{x} + \mu_{p}^{i - 1}(\mathbf{x}) | \mathbf{x}\in \mathbf{S}_t \}$, which is the reference shape wrapped with the mean deformation of the posterior model $\mathcal{GP}_{p}^{i-1}$. We then find the closest points $\mathbf{U}^{i}$ for each point in $\mathbf{S}_{reg}^{i}$ on $\mathbf{S}$, and update our posterior model as:
\begin{equation}
    \mathcal{GP}_{p}^{i+1}(\mu_{p}^{i+1}, k_{p}^{i+1}) = posterior(\mathcal{GP}_{p}^{0}, \mathbf{U}^{i} - \mathbf{S}_{reg}^{i})
\end{equation}
Since the raw scans in the MeIn3D database can be noisy, we exclude a pair of correspondence $(\mathbf{x}, \mathbf{U}(\mathbf{x}) )$ if $\mathbf{U}(\mathbf{x})$ is on the edge of $\mathbf{S}$ or the distance between $\mathbf{x}$ and $\mathbf{U}(\mathbf{x})$ exceed a threshold. After the final iteration we obtain the regression result $\mathbf{S}_{reg} = \{\mathbf{x} + \mu_{p}^{final}(\mathbf{x}) | \mathbf{x}\in \mathbf{S}_t \}$. We then non-rigidly align the face region of $\mathbf{S}_{reg}$ to the face region of the raw scan to obtain our final reconstruction.

In practice, we noticed that the reconstructions often produce unrealistic head shapes. We therefore modify the covariance matrix $\mathbf{K}_{U}$ before the Gaussian process regression. We first compute the principal components by decomposing $\mathbf{K}_{U}$, then reconstruct the covariance matrix using Eq. \ref{equ:compute_covariance} with fewer statistical components. With the full head reconstructions from the MeIn3D dataset, we then compute a new sample covariance matrix, and repeat the previous GP regression process to refine the reconstructions. Finally we perform PCA on the refined reconstructions to obtain our final refined model.
\section{Intrinsic evaluation of CFHM models}
\label{sec:experiments}
We name our combined full head model as the \emph{Combined Face \& Head  Model} (CFHM) and now show its comparative performance.
Following common practice, we evaluate our CFHM variations compared to LYHM by utilizing, \emph{compactness}, \emph{generalization} and \emph{specificity} \cite{davies2008statistical, brunton2014review, bolkart2015groupwise}. For all the subsequent experiments we utilise the original head scans of \cite{dai20173d} from which we have chosen 300 head meshes that were excluded from the training procedure. This test set was randomly chosen within demographic constrains to ensure ethnic, age and gender diversity. We name our model variations as: CFHM-reg built by the regression method, CFHM-GP built by the Gaussian processes kernels framework and finally, CFHM-ref built after refinement with Gaussian process regression. Also, we present bespoke modes in terms of age and ethnicity, constructed by the Gaussian processes kernels method coupled with refinement.

 The top graphs in Figure \ref{fig:intrinsic_exp} present the compactness measures of the CFHM models compared to LYHM. Compactness calculates the percentage of variance of the training data that is explained by the model, when certain number of principal components are retained. The models CFHM-reg, CFHM-GP express higher compactness compared to the model after the refinement. The compactness ability of the all proposed methods is far greater than the LYHM as it can be seen by the graph. Both global and bespoke CFHM models can be considered sufficiently compact.

 The center row of Fig. \ref{fig:intrinsic_exp} illustrates the generalization error which demonstrates the ability of the models to represent novel head shapes that are unseen during training. To compute the generalization error for a given number of principal components retained, we compute the per-vertex Euclidian distance between every sample of the test set and its corresponding model projection and then take the average value over all vertices and test samples. All of the proposed models exhibit far greater generalization capability compared to LYHM. The refined model CFHM-ref tends to generalize better than the other approaches, especially in the range of $20$ to $60$ components. Additionally, we plot the generalization error of the bespoke models against the CFHM-ref in center Figure \ref{fig:intrinsic_exp} (b). In order to derive a correct generalization measure for the bespoke CFHM-ref, for every mesh we use its demographic information, we project it on the subspace of the corresponding bespoke model and then we compute an overall average error. We observe that the CFHM-ref mostly outperforms the bespoke generalization models, which might be attributed to the fact that many of the specific models are trained from smaller cohorts, and so run out of interesting statistical variance.
 
Lastly, the bottom graphs of Figure \ref{fig:intrinsic_exp} show the specificity measures of the introduced models, which evaluate the validity of synthetic faces generated by a model. We randomly synthesize 5,000 faces from each model for a fixed number of components and measure how close they are to the real faces based on a standard per-vertex Euclidean distance metric. We observe that the model which holds the best error results is the proposed refined model CFHM-ref. The LYHM model demonstrates better specificity error than the  CFHM-reg, CFHM-GP models only in the first 20 components. Both of the proposed combined models exhibit steady error measures ($\approx 3.8$) after keeping components greater than 20. This is due to the higher compactness that both combined models demonstrate, which enables them to maintain certain specificity error after the 20 components. For all bespoke models we observe that the specificity errors attain particularly low values, in the range of $1$to $4$ mm. This is evidence that the synthetic faces generated by both global and bespoke CFHM models are realistic enough.

Our results show that our combination techniques yield models that are capable of exhibiting improved intrinsic characteristics compared to the original LYHM head model.

\begin{figure}[h]
    \centering
    \includegraphics[width=1\linewidth]{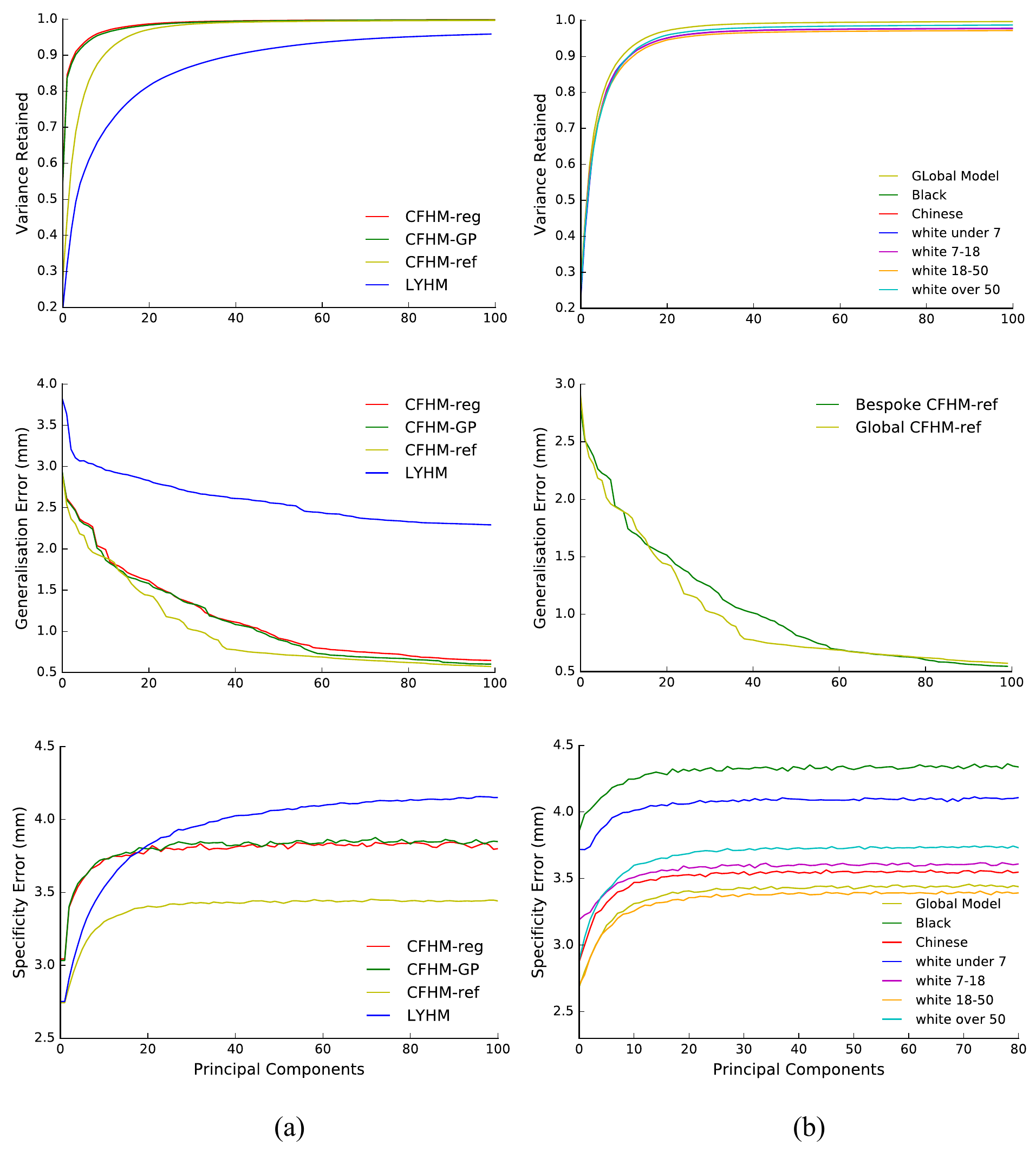}
    \caption{Characteristics of the CFHM models compared to LYHM. Top: compactness; Center: generalization; Bottom: specificity. Left column (a): different methods, Right column (b): demographic-specific 3DMMs based on the CFHM-ref model.}
    \label{fig:intrinsic_exp}
\end{figure}

\section{Head reconstruction from single images}
As an application experiment, we outline a methodology that enables us to reconstruct the entire head shape from unconstrained single images. We strictly utilize only one view/pose for head reconstruction in contrast to \cite{liang2016head} where multiple images were utilized with photometric constraints. We achieve this by regressing from a latent space that represents the 3D face and ear shape to the latent space of the full head models constructed by the proposed methodologies. We begin by building a PCA model of the inner face along with $50$ landmarks on each ear as described in \cite{zhou2017deformable}. We utilize the $10,000$ head meshes produced by our proposed methods. After building the face-ear PCA model, we project each one of the face-ear examples to get the associated shape parameters $\mathbf{p}_{e/f}$. Similarly, we project the full head mesh of the same identity to the full head PCA model in order to the acquire the latent shape parameters of the entire head $\mathbf{p}_h$. Similarly, as in section \ref{reg_combination}, we construct a regression matrix which works as a mapping from the latent space of the ear/face shape to the full head representation.

\begin{figure}[h]
    \centering
    \includegraphics[width=0.9\linewidth]{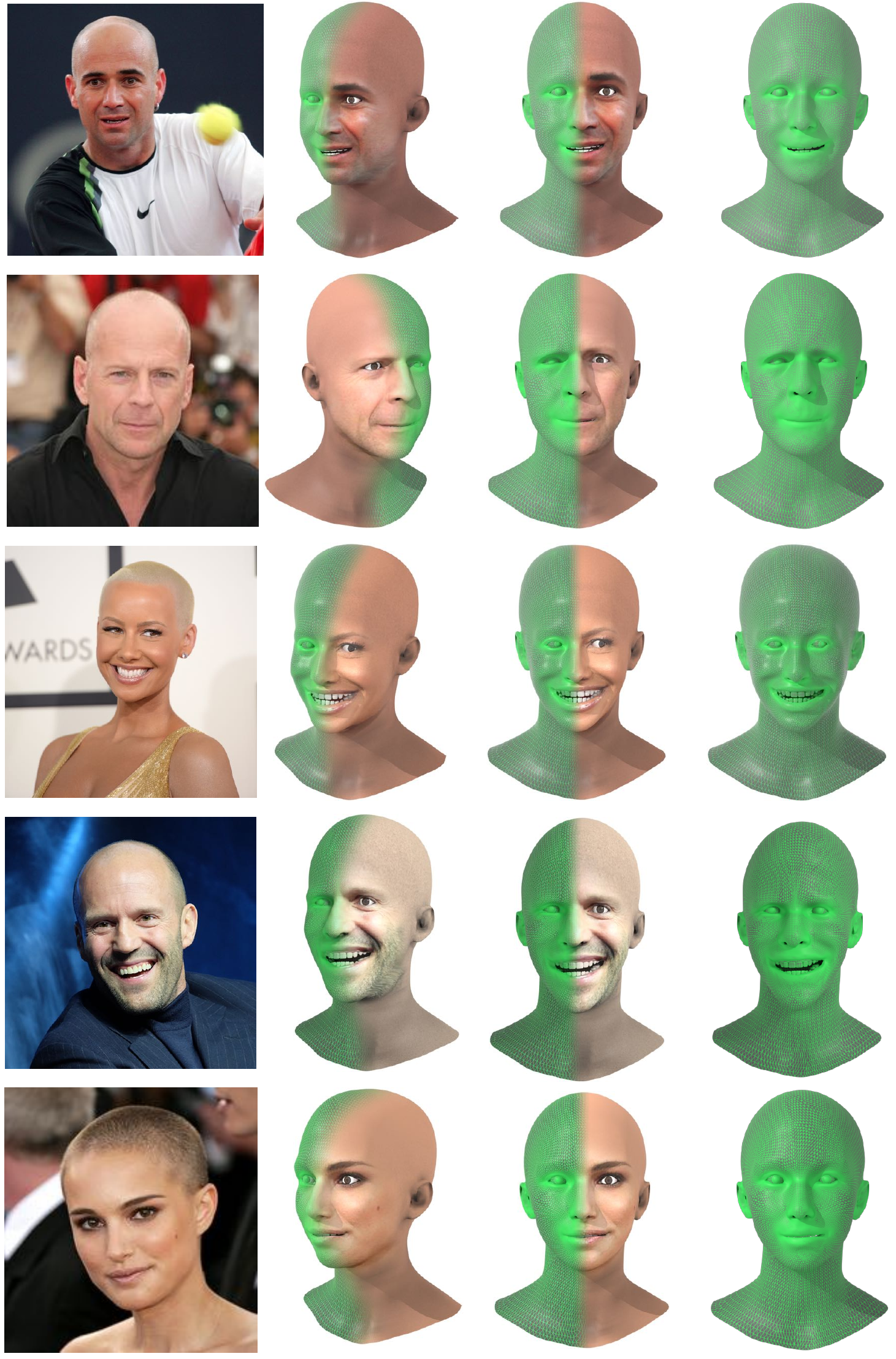}
    \caption{Qualitative results of our in-the-wild 3D head reconstruction. While the facial texture is reconstructed from the image domain, the eyes, the inner mouth and the head texture were created by an artist for a more realist representation.}
    \label{fig:heads_qual}
\end{figure}

In order to reconstruct the full head shape from 2D images we begin by fitting a face 3DMM utilizing the “In-the-Wild” feature-based texture algorithm proposed in \cite{booth20173d}. Afterwards, we implement an ear detector and an Active Appearance Model (AAM) as proposed in \cite{zhou2017deformable} to localize the ear landmarks in the 2D image domain. Since we have fitted a 3DMM in the image space, we already have the camera parameters,\ie, focal length, rotation, translation. To this effect, we can easily retrieve the ear landmarks in the 3D space by solving an inverse perspective-n-point problem \cite{lepetit2009epnp} given the camera parameters and the depth values of the fitted mesh. We mirror the 3D landmarks with respect to the z-axis to obtain the missing landmarks of the occluded ear. After acquiring the facial part and the ear landmarks we are able to attain the full head representation with the help of the regression matrix. Since each proposed method estimates a slightly different head shape for the $10,000$ face scans, we repeat the aforementioned procedure by building bespoke regression matrices for each head model. Some qualitative results can be seen in Figure \ref{fig:heads_qual}.

\begin{figure}[h]
    \centering
    \includegraphics[width=1\linewidth]{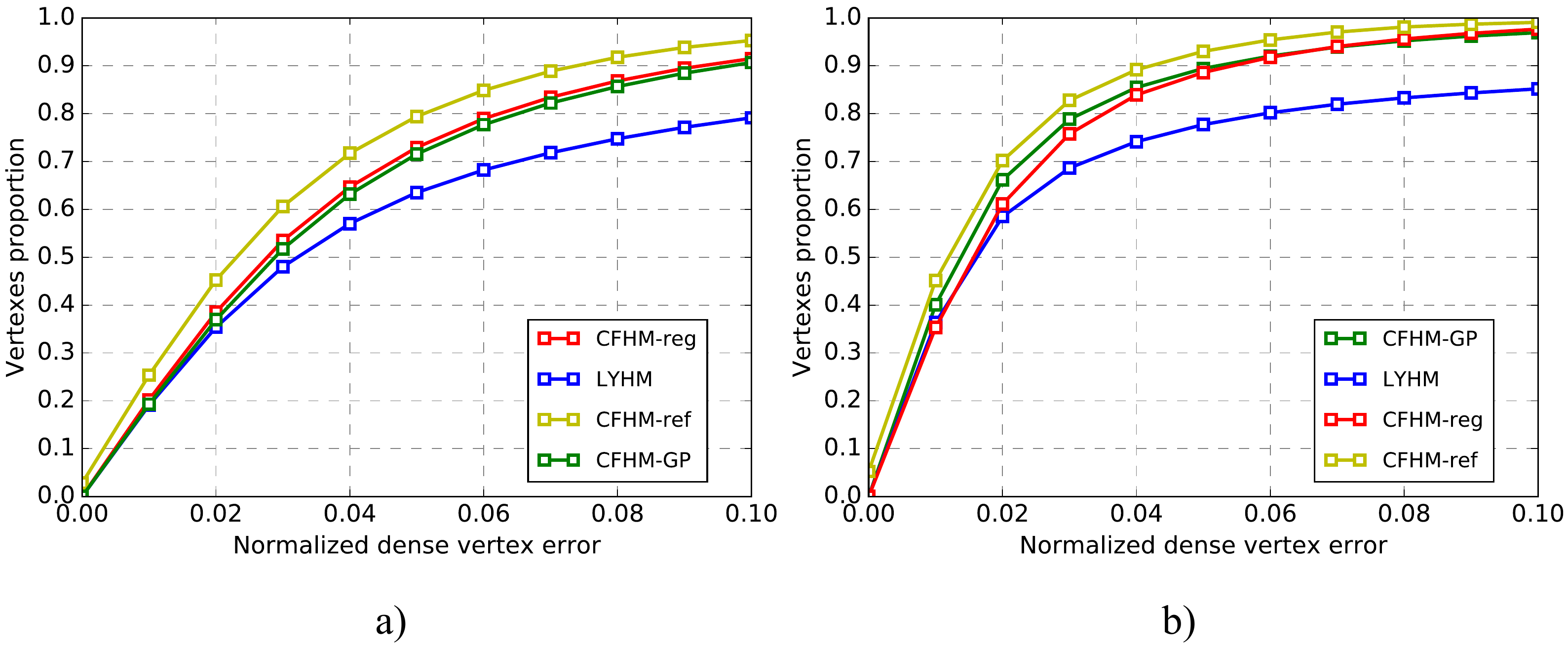}
    \caption{Accuracy results for head shape estimation, as cumulative error distributions of the normalized dense vertex errors. a) accuracy results based on the fitted facial meshes to rendered images, b) accuracy results based on the actual ground truth 3D facial meshes. Tables \ref{tab:dense_fit_error_1} and \ref{tab:dense_fit_error_2} report additional measures.}
    \label{fig:fit_exp}
\end{figure}

\begin{figure}[h]
    \centering
    \includegraphics[width=0.9\linewidth]{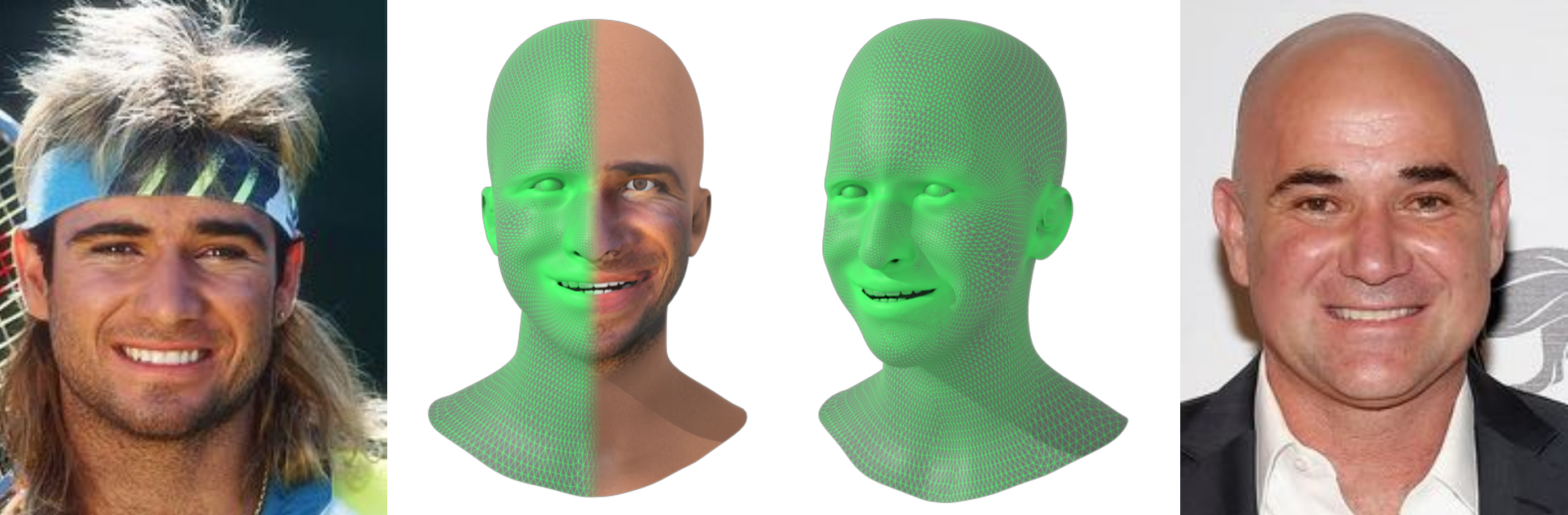}
    \caption{Regressing the full head when only face (left) is visible.}
    \label{fig:aggasi}
\end{figure}

We evaluate quantitatively our methodology by rendering $30$ distinct head scans from our test set in frontal and side poses varying from $20$ to $-20$ degrees around the $y$-axis in order for the ears to be visible in the image space. None of the head scans utilized for the evaluation, belong in the training process of any head model. We apply our previous procedure, where we fit a 3DMM face and we detect the ear landmarks in the image plane. Then for each method we exploit the bespoke regression matrix to predict the entire head shape. We measure the per-vertex error between the recovered head shape and the actual ground-truth head scan by projecting each point of the fitted mesh to the ground-truth and measuring the Euclidean distance. All distances are normalized based on the inter-ocular distance of the recovered mesh. Fig \ref{fig:fit_exp} shows the cumulative error distribution for this experiment, for the four models under test. Table \ref{tab:dense_fit_error_1} and \ref{tab:dense_fit_error_2} report the corresponding Area Under Curve (AUC) and failure rates for the fitted and the actual ground truth 3D facial meshes respectively. In both situations, the LYHM struggles to recover the head shapes. CFHM-reg and CFHM-GP perform equally, whereas the model after refinement attains the best results. Finally, Fig. \ref{fig:aggasi} illustrates regression of the full head shape, when only the face of the imaged subject is visible.

\begin{table}[!t]
\centering
\begin{tabular}{|l|cc|}
\hline
\emph{Method} & \emph{AUC} & \emph{Failure Rate (\%)} \\
\hline\hline
\textbf{CFHM-ref} & \textbf{0.751} & \textbf{3.64} \\
CFHM-reg & 0.693 & 6.88 \\
CFHM-GP & 0.681 & 7.55 \\
LYHM \cite{dai20173d}  & 0.605 & 19.21 \\
\hline
\end{tabular}
\vspace{0.2cm}
\caption{Head shape estimation accuracy results for the fitted facial meshes of our test set. Metrics are Area Under the Curve (AUC) and Failure Rate of the Cumulative Error Distributions of Fig.~\ref{fig:fit_exp}.}
\label{tab:dense_fit_error_1}
\end{table}

\begin{table}[!t]
\centering
\begin{tabular}{|l|cc|}
\hline
\emph{Method} & \emph{AUC} & \emph{Failure Rate (\%)} \\
\hline\hline
\textbf{CFHM-ref} & \textbf{0.880} & \textbf{0.62} \\
CFHM-GP & 0.844 & 2.46 \\
CFHM-reg & 0.831 & 1.69 \\
LYHM \cite{dai20173d}  & 0.739 & 14.10 \\
\hline
\end{tabular}
\vspace{0.2cm}
\caption{Head shape estimation accuracy results for the actual ground truth 3D facial meshes of our test set. Metrics are AUC and Failure Rate of the Cumulative Error Distributions of Fig.~\ref{fig:fit_exp}.}
\label{tab:dense_fit_error_2}
\end{table}

\section{Conclusion}
\label{sec:conclusions}

We presented a pipeline to fuse multiple 3DMMs into a single 3DMM and used it to combine the LSFM face model and the LYHM head model. 
This resulting 3DMM captured desirable properties of both constituent 3DMMs; namely high facial detail of the facial model and the full cranial shape variations of the head model. The augmented model is capable of representing and reconstructing any given face/head shape due to the high variation of facial and head appearances existing in the original models. We demonstrated that our methodology yielded a statistical model that is considerably superior to the original constituent models. Finally we illustrated the model's utility in full head reconstruction from a single image, where only the face is visible. In future work we will add a texture component to our models, extrapolating cranial texture from facial texture.


\newpage
{\small
\bibliographystyle{ieee}
\bibliography{egbib}
}
\end{document}